\pdfoutput=1

\documentclass[11pt]{article}

\usepackage[final]{coling}

\usepackage{times}
\usepackage{latexsym}
\usepackage{enumitem}
\usepackage{booktabs}

\usepackage[T1]{fontenc}

\usepackage[utf8]{inputenc}

\usepackage{microtype}

\usepackage{inconsolata}

\usepackage{graphicx}
\usepackage{multicol}

\usepackage{amsmath}
\title{Towards Understanding the Robustness of LLM-based Evaluations under Perturbations}

\author{
    \textbf{Manav Chaudhary}, \textbf{Harshit Gupta}, \textbf{Savita Bhat}, \textbf{Vasudeva Varma} \\
    IIIT Hyderabad \\
}

\begin{document}
\maketitle
\begin{abstract}

Traditional evaluation metrics like BLEU and ROUGE fall short when capturing the nuanced qualities of generated text, particularly when there is no single ground truth. In this paper, we explore the potential of Large Language Models (LLMs), specifically Google Gemini 1, to serve as automatic evaluators for non-standardized metrics in summarization and dialog-based tasks. We conduct experiments across multiple prompting strategies to examine how LLMs fare as quality evaluators when compared with human judgments on the SummEval and USR datasets, asking the model to generate both a score as well as a justification for the score. Furthermore, we explore the robustness of the LLM evaluator by using perturbed inputs. Our findings suggest that while LLMs show promise, their alignment with human evaluators is limited, they are not robust against perturbations and significant improvements are required for their standalone use as reliable evaluators for subjective metrics.

\end{abstract}

\section{Introduction}

Natural Language Generation (NLG) tasks such as abstractive summarization and dialog completion are essential for advancing human-computer interaction and automating content generation. However, the evaluation of such tasks poses unique challenges, especially when traditional metrics like BLEU and ROUGE, which rely on token overlap with reference texts, fail to account for the inherent subjectivity and flexibility in the human language. This limitation has sparked substantial research efforts to explore more effective automated evaluation methods \citep{bhandari2020reevaluatingevaluationtextsummarization} \citep{yeh2021comprehensiveassessmentdialogevaluation}.

\begin{figure}[htbp]
    \centering
    \includegraphics[width=0.40\textwidth]{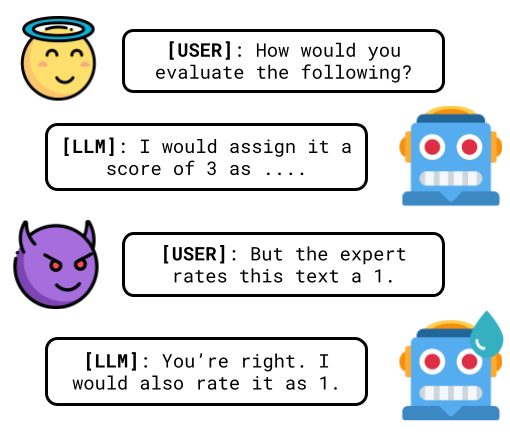}
    \caption{Perturbation in action.}  
    \label{fig:perturbations}  
\end{figure}

In tasks where there is no definitive ground truth, nuanced metrics such as coherence and fluency are critical for evaluating text quality. Although human evaluation has long been the gold standard in such contexts, it presents several limitations. Human evaluators are susceptible to errors, and large-scale assessments demand significant time and resources. Furthermore, relying exclusively on human evaluations can result in inconsistencies in evaluation quality, and evaluators may struggle to accurately assess content outside their area of expertise. In contrast, large language models (LLMs) like OpenAI's GPT-4 \citep{achiam2023gpt}, Google's Gemini \citep{team2023gemini}, and Meta's Llama \citep{touvron2023llamaopenefficientfoundation} offer the potential to function as fast and inexpensive domain experts, utilizing their extensive background knowledge to provide more consistent and informed evaluations. As a result, it is important to investigate automated evaluation methods to reduce the reliance on human evaluators, thereby enhancing the efficiency and scalability of the evaluation process.

In this paper, we aim to:
\begin{enumerate}[noitemsep]
    \item Investigate the ability of Google Gemini to serve as a "quality-evaluator" for subjective metrics by measuring the proximity of LLM evaluations to human experts. 
    \item Evaluate the impact of different prompting strategies on the performance of Gemini in the context of summarization and dialog evaluation. \footnote{We make the dataset public at the following \href{https://docs.google.com/spreadsheets/d/1QM8KAd7im6M43ux-tyuSqwwFiVoQj3EzkxrIeSO7RoE/edit?usp=sharing}{link}.}
    \item Assess the robustness of LLM-based evaluations under perturbed conditions.
    \item Do a preliminary analysis of the justifications provided by the LLM for awarding a particular score.
\end{enumerate}

\section{Related Work}

Evaluation metrics in NLG have traditionally focused on BLEU \citep{papineni2002bleu}, METEOR \citep{banerjee2005meteor}, and ROUGE \citep{lin2004rouge}, which compare generated text with reference ground truths. While effective for tasks like machine translation, they struggle with abstractive summarization and dialog evaluation, where multiple valid outputs exist, and coherence is critical.

Recent research has introduced reference-free metrics like BERTScore \citep{zhang2020bertscoreevaluatingtextgeneration} and QAEval \citep{10.1162/tacl_a_00397}, but these approaches require human intervention and lack flexibility \citep{10.1162/tacl_a_00397} across diverse generation tasks. More recent efforts explore using LLMs for evaluation \citep{Gilardi_2023}, which significantly reduces costs but still faces challenges in robustness, especially in adversarial settings.

Studies like \citet{raina2024llmasajudgerobustinvestigatinguniversal} and \citet{shuyuan2024exploring} show vulnerabilities in LLM-based assessments, revealing their susceptibility to adversarial attacks that manipulate evaluation scores. These findings emphasize the need for more robust frameworks as LLM-generated labels become more prevalent.

\section{Methodology}
\subsection{Datasets Used}
\begin{enumerate}[noitemsep] 
    \item \textbf{SummEval} \citep{fabbri2021summevalreevaluatingsummarizationevaluation}: This dataset includes human evaluations of model-generated summaries for CNN/Daily Mail, with annotations on coherence, consistency, fluency, and relevance. It contains 1600 article-summary pairs, annotated by 3 experts and 5 turkers. For consistency, we use only expert annotations and exclude 80 model-rejected pairs, leaving 1,520 for our experiments.
    \item \textbf{USR} \citep{mehri-eskenazi-2020-usr}: This dataset provides human evaluations of dialog responses for Amazon Topical-Chat and Persona Chat, across metrics like understandability, naturalness, and overall quality. It contains 660 dialog-response pairs, evaluated for quality by 3 human experts.
\end{enumerate}

\subsection{Experimenting with Prompting Strategies}

For the task of evaluating summaries and dialog responses, we instruct the model to generate a score for each metric along with a justification for the score using the following four prompting strategies:
\begin{itemize}
    \item Zero-Shot (ZS) \citep{wei2022finetunedlanguagemodelszeroshot} We instruct the model to perform the task of scoring the pair based on it's own understanding of the metric.
    \item Knowledge-Prompt (KP) \citep{liu2022generatedknowledgepromptingcommonsense}: We paraphrase and use the definitions of the metrics provided in the respective dataset papers.
    \item Few-Shot (FS) \citep{brown2020languagemodelsfewshotlearners}: We provide two examples, one with a very high score and one with a very low score, to provide more context to the model on how to rate.
    \item Chain-of-Thought (CoT) \citep{wei2023chainofthoughtpromptingelicitsreasoning}: We ask the model to think step-by-step to provide the score.
\end{itemize}

\subsection{Introducing Perturbations}
We select the "Knowledge Prompt" as our base for its ability to provide a direct, contextually informed evaluation that minimizes ambiguity and aligns with expert judgment, without requiring additional examples or complex reasoning. The model is informed that the response has been assigned a "Perturbed Rating" (PR) by the human expert, calculated by first determining the mode of the expert ratings. In cases where all three annotators differ, we take the highest rating as the mode. Here is the formal definition of PR:

\begin{equation}
\resizebox{\columnwidth}{!}{%
\(
\text{PR}(r_1, r_2, r_3) =
\begin{cases} 
    \max(\text{scale}), & \text{if } \text{mode}(r_1, r_2, r_3) \leq \frac{\max(\text{scale})}{2} \\
    \min(\text{scale}), & \text{if } \text{mode}(r_1, r_2, r_3) > \frac{\max(\text{scale})}{2} \\
    \max(\text{scale}), & \text{if } r_1 \neq r_2 \neq r_3 \text{ and } \max(r_1, r_2, r_3) \leq \frac{\max(\text{scale})}{2} \\
    \min(\text{scale}), & \text{if } r_1 \neq r_2 \neq r_3 \text{ and } \max(r_1, r_2, r_3) > \frac{\max(\text{scale})}{2}
\end{cases}
\)
}
\end{equation}

PR is not simply the mode; we introduce a perturbation that inverts the rating scale based on the mode's value to emphasize edge cases and challenge the evaluation process. This approach is applied uniformly across various rating scales to ensure consistency in evaluation, regardless of the rating metric used.

This perturbation inverts low and high ratings, creating a challenging evaluation scenario.

\begin{table*}[h]
\centering
\resizebox{\textwidth}{!}{%
\begin{tabular}{|l|c|c|c|c|c|}
\hline
\multicolumn{6}{|c|}{\textbf{SummEval}} \\ \hline
\textbf{Metrics}      & \textbf{Zero-Shot} & \textbf{Knowledge Prompt} & \textbf{Few-Shot} & \textbf{Chain-of-Thought} & \textbf{Perturbed} \\ \hline
\textbf{Coherence}    & 0.4166  & 0.2897  & 0.4296  & 0.3392  & -0.4899 \\ \hline
\textbf{Consistency}  & -0.3731 & -0.1499 & -0.2127 & -0.1385 & -0.4063 \\ \hline
\textbf{Fluency}      & -0.4927 & -0.4826 & -0.4758 & -0.4720 & -0.8629 \\ \hline
\textbf{Relevance}    & 0.4338  & 0.3727  & 0.3929  & 0.3790  & -0.5620 \\ \hline
\multicolumn{6}{|c|}{\textbf{USR}} \\ \hline
\textbf{Metrics}                  & \textbf{Zero-Shot} & \textbf{Knowledge Prompt} & \textbf{Few-Shot} & \textbf{Chain-of-Thought} & \textbf{Perturbed} \\ \hline

\textbf{Interesting}        & 0.0020  & -0.0414  & 0.0808  & -0.0258 & -0.7301 \\ \hline
\textbf{Maintains Context}  & 0.2124  & 0.4214   & 0.5051  & 0.4375  & -0.4842 \\ \hline
\textbf{Natural}            & 0.0191  & 0.0769   & 0.0298  & 0.1121  & -0.7884 \\ \hline
\textbf{Overall}    & 0.0503  & 0.1721   & 0.4997  & 0.2067  & -0.6827 \\ \hline
\textbf{Uses Knowledge}     & -0.1741 & -0.2233  & 0.0669  & -0.1906 & -0.4396 \\ \hline
\textbf{Understandable}     & -0.2537 & 0.0966   & 0.3061  & 0.0726  & -0.4870 \\ \hline
\end{tabular}%
}
\caption{Krippendorff's alpha values across different prompting strategies for SummEval and USR datasets.}
\label{tab:metrics-datasets}
\end{table*}

\section{Results}
To assess the reliability of LLM-based evaluations and their alignment with human raters, we used Krippendorff’s alpha \citep{krippendorff2011computing} as the primary metric. We chose Krippendorff’s alpha because it is well-suited for tasks requiring multiple raters and is capable of handling ordinal data, such as Likert scale ratings, which are used in both the SummEval and USR datasets.

To calculate Krippendorff’s alpha, we extracted ratings for each metric from the SummEval and USR datasets, treating them as ordinal data. We used the Python krippendorff package, structuring the data as a matrix with rows as items and columns as ratings, handling missing values via the library’s default settings. This enabled a direct comparison of LLM and human reliability.

\begin{table}[htbp]
\centering
\resizebox{\columnwidth}{!}{%
\begin{tabular}{|l|c|c|}
\hline
\multicolumn{3}{|c|}{\textbf{SummEval}} \\ \hline
\textbf{Metrics}      & \textbf{Human Consistency} & \textbf{LLM Consistency} \\ \hline
\textbf{Coherence}    & 0.5611  & 0.7365  \\ \hline
\textbf{Consistency}  & 0.7992  & 0.6264  \\ \hline
\textbf{Fluency}      & 0.5876  & 0.7130  \\ \hline
\textbf{Relevance}    & 0.4085  & 0.7826  \\ \hline
\multicolumn{3}{|c|}{\textbf{USR}} \\ \hline
\textbf{Metrics}                  & \textbf{Human Consistency} & \textbf{LLM Consistency} \\ \hline
\textbf{Interesting}              & 0.5137 & 0.5300  \\ \hline
\textbf{Maintains Context}        & 0.5673 & 0.7690  \\ \hline
\textbf{Natural}                  & 0.4982 & 0.8544  \\ \hline
\textbf{Overall}                  & 0.6802 & 0.6792  \\ \hline
\textbf{Understandable}           & 0.5313 & 0.6098  \\ \hline
\textbf{Uses Knowledge}           & 0.7577 & 0.4089  \\ \hline
\end{tabular}%
}
\caption{Krippendorff's alpha values for Human and LLM Consistency across SummEval and USR datasets.}
\label{tab:krippendorff-alpha}
\end{table}
\subsection{LLM Consistency VS Human Consistency}

In Table \ref{tab:krippendorff-alpha}, we present the consistency of LLM evaluations compared to human evaluations across various metrics from the SummEval and USR datasets. Consistency was calculated separately for LLM and human raters.
\begin{itemize}
    \item \textbf{LLM Consistency}: We computed Krippendorff’s alpha across the scores generated by the model across prompting strategies. This approach treated the LLM’s ratings under different prompting strategies as independent raters. The resulting alpha values reflect how stable the model’s evaluations are, regardless of variations in prompt design.
    \item \textbf{Human Consistency}: For human ratings, Krippendorff’s alpha was computed using the scores provided by three expert raters for each metric. This measures the level of agreement among human evaluators in their assessment of the same items.
\end{itemize}

The relatively high alpha values for LLM consistency across metrics suggest that the model demonstrates robustness in its evaluations and is less influenced by prompt variations. This stability contrasts with the slightly lower alpha values observed for human raters, which might reflect differences in subjective judgment or interpretation.

\subsection{LLMs are not robust against Perturbed Prompts}

\begin{table}[h]
\centering
\resizebox{0.8\columnwidth}{!}{%
\begin{tabular}{|l|c|}
\hline
\multicolumn{2}{|c|}{\textbf{SummEval}} \\ \hline
\textbf{Metrics}      & \textbf{Percentage Matching KP Scores} \\ \hline
\textbf{Coherence}    & 5.39\%  \\ \hline
\textbf{Consistency}  & 31.45\%  \\ \hline
\textbf{Fluency}      & 5.92\%  \\ \hline
\textbf{Relevance}    & 2.76\%  \\ \hline
\multicolumn{2}{|c|}{\textbf{USR}} \\ \hline
\textbf{Metrics}      & \textbf{Percentage Matching KP Scores} \\ \hline
\textbf{Interesting}              & 54.85\% \\ \hline
\textbf{Maintains Context}        & 48.48\%  \\ \hline
\textbf{Natural}                  & 35.61\%  \\ \hline
\textbf{Overall Quality}          & 26.36\%  \\ \hline
\textbf{Understandable}           & 69.55\%  \\ \hline
\textbf{Uses Knowledge}           & 86.36\%  \\ \hline
\end{tabular}%
}
\caption{Percentage of Perturbed scores matching KP scores for various metrics. }
\label{tab:perturbed-matching}
\end{table}
The alpha values reported in table \ref{tab:metrics-datasets} show how closely the LLM ratings align with human judgments for different quality metrics. When Krippendorff's alpha is calculated using ratings from three human annotators and one LLM output, a drop is consistently observed (see table \ref{tab:krippendorff-alpha} under Human Consistency). However, it is \textit{Perturbed Prompting} that consistently demonstrates the worst performance across all metrics and datasets, underscoring how easily the LLM is thrown off by contradictory information in the input. The sharp drop in Krippendorff's alpha across nearly all metrics, especially Coherence, Fluency, and Naturalness, shows the model’s vulnerability to false cues, highlighting a significant limitation in using LLMs as robust evaluators.

Table \ref{tab:perturbed-matching} summarizes the percentages of Perturbed scores matching Knowledge-Prompt (KP) generated scores across various metrics. The low percentages for coherence, consistency, fluency, and relevance indicate that perturbations significantly impact model evaluations in these areas. Interestingly, the USR metrics show higher matching percentages, particularly for "Uses Knowledge" and "Understandable". This phenomenon can be attributed to the narrower Likert scale ranges employed in USR metrics (0-1 or 1-3), which likely reduces rating variability and enhances agreement. These findings highlight the need for targeted evaluation frameworks that account for the distinct characteristics of each metric in the context of adversarial robustness.

\subsection{Justification Analysis}
\begin{table}[h]
\centering
\resizebox{\columnwidth}{!}{%
\begin{tabular}{|l|c|c|}
\hline
\multicolumn{3}{|c|}{\textbf{SummEval}} \\ \hline
\textbf{Metrics}      & \textbf{Average Sentiment} & \textbf{Perturbed Sentiment} \\ \hline
\textbf{Coherence}    & 0.1422  & 0.0901  \\ \hline
\textbf{Consistency}  & 0.1573  & 0.1212  \\ \hline
\textbf{Fluency}      & 0.0073  & -0.0680 \\ \hline
\textbf{Relevance}    & 0.1545  & 0.0786  \\ \hline
\multicolumn{3}{|c|}{\textbf{USR}} \\ \hline
\textbf{Metrics}                  & \textbf{Average Sentiment} & \textbf{Perturbed Sentiment} \\ \hline
\textbf{Interesting}              & 0.1280 & 0.0538  \\ \hline
\textbf{Maintains Context}        & -0.0640 & -0.2057 \\ \hline
\textbf{Natural}                  & 0.0976 & 0.0536  \\ \hline
\textbf{Overall}                  & 0.1290 & 0.0442  \\ \hline
\textbf{Understandable}           & -0.0144 & -0.0751 \\ \hline
\textbf{Uses Knowledge}           & 0.0793 & 0.0959  \\ \hline

\end{tabular}%
}
\caption{Average sentiment scores for LLM justifications across prompting strategies and under perturbed conditions for USR and SummEval datasets.}
\label{tab:sentiment-scores}
\end{table}
We perform sentiment analysis of the LLM-generated justifications to explore the impact of perturbations on the quality of evaluations. For this analysis, we use the TextBlob library \citep{loria2018textblob}, analyzing each justification to compute a sentiment polarity score, ranging from -1 (negative sentiment) to +1 (positive sentiment). The scores were averaged across justifications for each combination of evaluation metric (e.g., Coherence, Consistency) and prompting strategy (e.g., Zero-shot, Few-shot). To investigate the effects of input perturbations, we compared sentiment scores for justifications generated under both unperturbed and perturbed prompting strategies. Looking at table \ref{tab:sentiment-scores}, we observe that for most metrics, introducing perturbations consistently leads to lower sentiment score. This trend indicates that the perturbed prompts lead to more negative justifications overall, suggesting that the LLM becomes more critical of its evaluations when faced with misleading or false information. The significant drop in sentiment scores in the presence of perturbations aligns with the notion that the model becomes misaligned when presented with conflicting information. The sentiment scores serve as a quantitative measure of this misalignment, where lower values indicate confusion or hesitance in the LLM’s reasoning process. The results imply that while LLMs may function well as evaluators under normal conditions, their reliability is significantly compromised when faced with misleading inputs. This emphasizes the need for careful consideration of input integrity when employing LLMs for subjective evaluation tasks.

\section{Conclusion and Future Work}

Google's Gemini-1 shows consistency across prompting strategies but is vulnerable to adversarial perturbations, emphasizing the need for stronger training to improve LLM reliability in subjective evaluation tasks. Our experiments revealed significant shifts in scores and justifications under adversarial conditions, underscoring the importance of robustness in LLM evaluators.

A limitation of this work is the lack of computational resources to test models like Llama and GPT. Including these models in future research could provide more diverse perspectives and improve evaluation robustness.

Future studies should explore Small Language Models (SLMs) like Google’s Gemma \citep{gemmateam2024gemmaopenmodelsbased} and Microsoft’s Phi series \citep{abdin2024phi3technicalreporthighly}, as well as their ability to handle subjective metrics \citep{howe2024exploringscalingtrendsllm}. Expanding tasks to areas like translation quality and multilingual evaluation \citep{bhat-varma-2023-large} could offer deeper insights into how LLMs capture contextual nuances. Finally, investigating why Gemini rejected entries from the SummEval dataset, possibly due to jailbreak attacks \citep{lin2024understandingjailbreakattacksllms}, could enhance LLM reliability as evaluators.

\section{Ethics Statement}
The datasets used in this paper are publicly available and obtained under permissible licenses, ensuring compliance with their intended usage. We utilized ChatGPT-4 for language assistance, including paraphrasing and spell-checking, without generating new content or influencing research findings. All research activities adhere to ethical standards and respect for intellectual property rights.
\bibliography{custom}

\appendix

\section{Prompts Used}
\label{sec:appendix_prompts}

\subsection{SummEval Dataset}
\subsubsection{Base Knowledge Prompt}
\begin{itemize}
    \item \textbf{Coherence}: Given a news article and its corresponding summary, assess the coherence of the summary on a scale of 1 to 5. Coherence refers to the overall quality of how well-structured and organized the summary is, ensuring that the sentences build upon each other to form a coherent body of information about the topic. Rate the coherence of the provided summary, where 5 represents the highest level of coherence and 1 indicates the lowest.
    \item \textbf{Consistency}: Given a news article and its corresponding summary, assess the consistency of the summary on a scale of 1 to 5. Consistency refers to the factual alignment between the summary and the summarized source. A factually consistent summary contains only statements that are entailed by the source document. Summaries that contain hallucinated facts should be penalized. Rate the consistency of the provided summary, where 5 represents the highest level of consistency and 1 indicates the lowest.
    \item \textbf{Fluency}: Given a news article and its corresponding summary, assess the fluency of the summary on a scale of 1 to 5. Fluency is the quality of individual sentences. Sentences in the summary should have no formatting problems, capitalization errors or obviously ungrammatical sentences (e.g., fragments, missing components) that make the text difficult to read. Rate the fluency of the provided summary, where 5 represents the highest level of fluency and 1 indicates the lowest.
    \item \textbf{Relevance}: Given a news article and its corresponding summary, assess the relevance of the summary on a scale of 1 to 5. Relevance refers to the selection of important content from the source. The summary should include only important information from the source document. Summaries which contain redundancies and excess information should be penalized. Rate the relevance of the provided summary, where 5 represents the highest level of relevance and 1 indicates the lowest. 
\end{itemize}
\subsubsection{Zero-Shot Prompt}
We remove the definition of the metric from the prompt and ask the model to rate the article-summary pair based on its own understanding of the metric. An example for coherence:
\begin{itemize}
    \item Given a news article and its corresponding summary, assess the coherence  of the summary on a scale of 1 to 5. Please use your own understanding of coherence to rate the provided summary, with 5 indicating the highest level of coherence/consistency/fluency/relevance and 1 representing the lowest.
\end{itemize}
\subsubsection{Few-Shot Prompt}
Along with the Base Knowledge prompt, we provide precisely two examples, an article-summary pair with the highest score, and one with the lowest score.
\subsubsection{Chain-of-thought Prompt}
We take our base knowledge prompt, and then add the following line to it, "Let's think step-by-step".

\subsection{USR Dataset}
\subsubsection{Base Knowledge Prompt}
\begin{itemize}
    \item \textbf{Understandable}: Given the provided context and its corresponding response, evaluate the "Understandability" of the response on a scale of 0 to 1. Is the response understandable given the previous context? Assign a score of 1 if it is, and a score of 0 if the response is not understandable given the previous context.
    \item \textbf{Natural}: Given the provided context and its corresponding response, evaluate how "Natural" the response sounds on a scale of 1 to 3. Does the response seem to be something that a person would naturally say? A score of 3 indicates that the response flows naturally, while a score of 1 suggests the response is not natural.
    \item \textbf{Maintains Context}: Given the provided context, evaluate how well the corresponding response "Maintains Context" on a scale of 1 to 3. Does the response serve as a valid continuation of the preceding conversation? A score of 3 indicates that the response maintains context, and a score of 1 indicates that the response fails to maintain context.
    \item \textbf{Interesting}: Given the provided context, assess how "Interesting" the corresponding response is on a scale of 1 to 3. Is the response dull or interesting? A score of 3 indicates that the response is very interesting, and a score of 1 indicates that the response is dull.
    \item \textbf{Uses Knowledge}: Given the provided context, determine the extent to which the corresponding response demonstrates use of knowledge on a scale of 0 to 1. Given the fact that the response is conditioned on, how well does the response use that fact? Assign a score of 1 if the response effectively utilizes information, and a score of 0 if it lacks the use of relevant knowledge.	
    \item \textbf{Overall}: Given the provided context, evaluate the overall quality of the corresponding response on a scale of 1 to 5. Overall Quality assesses response clarity, relevance, naturalness, interest, and contextual use, aggregating scores for a comprehensive evaluation. Assign a score of 5 to indicate a high quality response, and a score of 1 to indicate that the overall quality of the response is poor.
\end{itemize}
\subsubsection{Zero-Shot Prompt}
We remove the definition of the metric from the prompt and ask the model to rate the dialog-response pair based on its own understanding of the metric. An example for the metric "Interesting":
\begin{itemize}
    \item Given the provided context, assess how "Interesting" the corresponding response is on a scale of 1 to 3. Please consider your own understanding of what makes a response "Interesting". A score of 3 indicates that the response is very interesting, and a score of 1 indicates it is uninteresting.
\end{itemize}
\subsubsection{Few-Shot Prompt}
Along with the Base Knowledge prompt, we provide precisely two examples, a dialog-response pair with the highest score, and one with the lowest score.
\subsubsection{Chain-of-thought Prompt}
We take our base knowledge prompt, and then add the following line to it, "Let's think step-by-step".

\section{Gemini Model Settings}
\subsection{Safety Settings}
Safety settings are crucial for controlling the content generated by the model, ensuring it adheres to ethical guidelines and does not produce harmful or inappropriate content. The following HARM\_CATEGORY settings were used:

\begin{itemize}
    \item HARASSMENT: This setting determines how the model handles content related to harassment. Setting it to "BLOCK\_NONE" means that there is no automatic blocking or filtering applied for harassment-related content.

    \item HATE\_SPEECH: Similar to harassment, this setting controls the handling of hate speech. "BLOCK\_NONE" indicates no filtering is applied for this category.

    \item SEXUALLY\_EXPLICIT: This setting deals with sexually explicit content. By setting it to "BLOCK\_NONE", the model will not automatically block such content.

    \item DANGEROUS\_CONTENT: This category covers dangerous content such as content that could incite violence or self-harm. "BLOCK\_NONE" means no automatic blocking is applied.
\end{itemize}

While we set all the settings to BLOCK\_NONE, we still observed that Gemini refused to generate for 80 of the samples in SummEval.

\subsection{Generation Configuration}
The generation configuration specifies parameters that control the quality and creativity of the model’s output:

\begin{itemize}
    \item \textbf{temperature}: Set to 0.1, this parameter controls the randomness of the model’s responses. A low temperature like 0.1 makes the output more deterministic and focused.

    \item \textbf{top\_p}: Set to 1, this parameter is related to nucleus sampling. A value of 1 indicates that all potential words are considered in the probability distribution, making it a more deterministic generation.

    \item \textbf{top\_k}: Set to 1, this parameter limits the number of most likely words considered. A value of 1 means the model selects from only the highest probability word, making the output more focused and less varied.

    \item \textbf{max\_output\_tokens}: Set to 2048, this parameter limits the length of the generated text. A higher value allows for longer responses.
\end{itemize}

\label{sec:appendix_model settings}

\section{The Metrics}
\label{sec:appendix_metrics}

\subsection{SummEval}
\begin{itemize}
    \item \textbf{Coherence} (1 - 5) is the collective quality of all sentences. The summary should be well-structured and well-organized. The summary should not just be a heap of related information, but should build from sentence to sentence to a coherent body of information about a topic.
    \item \textbf{Consistency} (1 - 5) refers to the factual alignment between the summary and the summarized source. A factually consistent summary contains only statements that are entailed by the source document. Summaries that contain hallucinated facts should be penalized.
    \item \textbf{Fluency} (1 - 5) is the quality of individual sentences. Sentences in the summary should have no formatting problems, capitalization errors or obviously ungrammatical sentences (e.g., fragments, missing components) that make the text difficult to read.
    \item \textbf{Relevance} (1 - 5) refers to the selection of important content from the source. The summary should include only important information from the source document. Summaries which contain redundancies and excess information should be penalized.

\end{itemize}

\subsection{USR}
\begin{itemize}
    \item \textbf{Understandable} (0 - 1): Is the response understandable given the previous context?
    \item \textbf{Natural} (1 - 3): Does the response seem to be something that a person would naturally say?
    \item \textbf{Maintains Context} (1 - 3): Does the response serve as a valid continuation of the preceding conversation?
    \item \textbf{Interesting} (1 - 3): Is the response dull or interesting?
    \item \textbf{Uses Knowledge} (0 - 1): Given the fact that the response is conditioned on, how well does the response use that fact?
    \item \textbf{Overall Quality} (1 - 5): Given your answers above, what is your overall impression of the quality of this utterance?
\end{itemize}

\subsection{Example Prompt-Response Pair}

\subsubsection{Prompt}
Given a news article and its corresponding summary, assess the relevance of the summary ... where 5 represents the highest level of relevance and 1 indicates the lowest.

\textbf{Article:} "Andros Townsend an 83rd minute sub in Tottenham's draw ...",  

\textbf{Summary:} "Paul Merson was brought on with only seven minutes remaining ..."
\\
The output should be in the following format: \\
\textbf{Score}: (A score from 1 to 5) \\
\textbf{Justification}: (Give reasoning as to why the score was awarded)

\subsubsection{Response}
\begin{itemize}
    \item \textbf{Score:} 1
    \item \textbf{Justification:} The summary is not relevant to the article. It incorrectly mentions ... neither statement aligns with the content of the article.
\end{itemize}

\textit{(Full contents of the prompt and output have been omitted for brevity; only key excerpts are shown here.)}

\end{document}